\documentclass{article}
\usepackage{spconf,amsmath,amssymb,graphicx,booktabs,enumitem,bm}
\renewcommand{\baselinestretch}{0.94}
\usepackage{pgfplots,tikzscale}

\def\distfunc{\mathrm{dist}}
\def\simfunc{\mathrm{sim}}

\title{DEEP ONE-CLASS CLASSIFICATION USING INTRA-CLASS SPLITTING}

\name{Patrick Schlachter, Yiwen Liao and Bin Yang}
\address{Institute of Signal Processing and System Theory, University of Stuttgart, Germany}

\begin{document}
\maketitle

\begin{abstract}
This paper introduces a generic method which enables to use conventional deep neural networks as end-to-end one-class classifiers. The method is based on splitting given data from one class into two subsets. In one-class classification, only samples of one normal class are available for training. During inference, a closed and tight decision boundary around the training samples is sought which conventional binary or multi-class neural networks are not able to provide. By splitting data into typical and atypical normal subsets, the proposed method can use a binary loss and defines an auxiliary subnetwork for distance constraints in the latent space. Various experiments on three well-known image datasets showed the effectiveness of the proposed method which outperformed seven baselines and had a better or comparable performance to the state-of-the-art.
\end{abstract}


\section{Introduction}
\label{sec:intro}
% one-class classification: definition and motivation
One-class classification describes special classification problems in which only samples from one class, the so-called normal class, are available for training. During inference, the task is to discriminate normal samples from samples of the other class, the so-called abnormal or anomaly class.

% problem: state-of-the-art one-class classifiers have limited performance on high-dimensional data
Conventional one-class classifiers such as the one-class support vector machine (OCSVM)~\cite{ocsvm2001} or the support vector data description (SVDD)~\cite{Tax2004} have limited performance on complex raw data such as natural images because of their sensitive hyperparameters $\gamma$, $\nu$ and $C$. Furthermore, they require hand-crafted features which are task-dependent and have to be carefully figured out by experts.

% deep learning
In contrast to traditional machine learning methods, deep learning can benefit from a huge amount of data and achieves better performances in complex tasks such as image classification, natural language processing and speech recognition~\cite{deeplearning2016}. Accordingly, an obvious research direction is to use deep learning methods for one-class classification. 

\begin{figure}[t]
	\centering
	\begin{minipage}[b]{0.48\linewidth}
		\centering
		\centerline{\includegraphics[width=\linewidth]{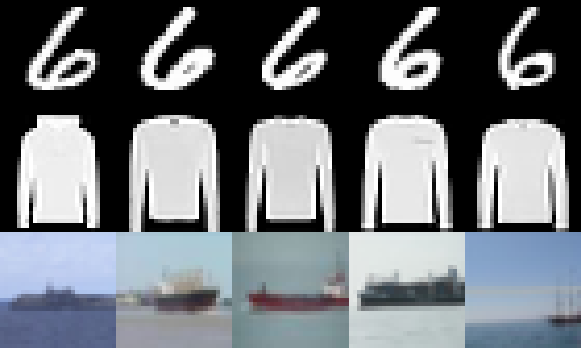}}
		\centerline{(a) Typical examples}\medskip
	\end{minipage}
	\hfill
	\begin{minipage}[b]{.48\linewidth}
		\centering
		\centerline{\includegraphics[width=\linewidth]{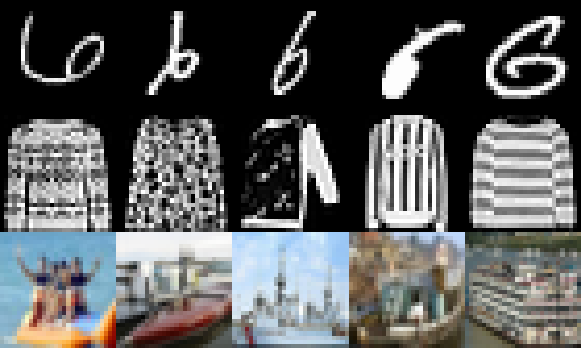}}
		\centerline{(b) Atypical examples}\medskip
	\end{minipage}
	\vspace{-0.4cm}
	\caption{Examples for typical and atypical samples.}
	\label{fig:examples_typical_atypical}
	\vspace{-0.3cm}
\end{figure}

\begin{figure*}[htb]
	\centering
	\includegraphics{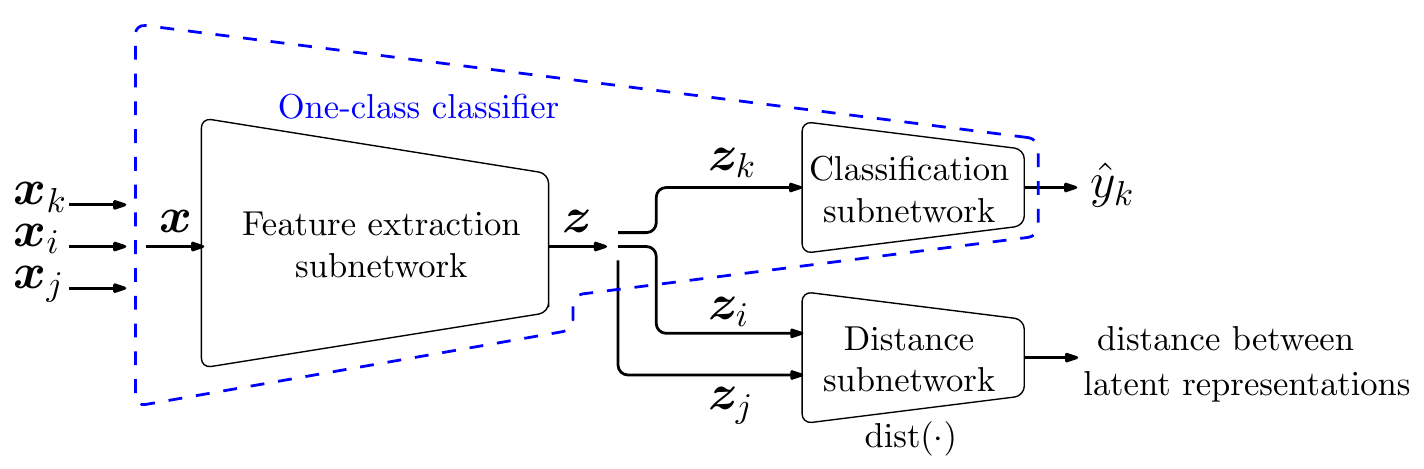}
	\vspace{-0.1cm}
	\caption{The architecture of the proposed method.}
	\label{fig:network}
\end{figure*}

% literature: existing solutions + problems
Indeed, there exist only few deep learning approaches to one-class classification. One typical method is to train an autoencoder by normal samples only and to use the reconstruction error as an indication of their class affiliation~\cite{sakurada2014anomaly}. Various research was conducted in this field. For instance, recent literature includes methods based on variational autoencoders~\cite{an2015variational,Kawachi2018}, additional regularization terms to a mean squared error (MSE) cost function for higher robustness~\cite{zhou2017anomaly} and a combination of clustering and constraints on the latent space of an autoencoder~\cite{aytekin2018clustering}. However, all these error-based methods have limited performance for complex datasets, because the pixel-wise error does not always match the human understanding~\cite{hog2005}. Hence, either the given normal samples should be distributed closely to each other in the original data space or feature engineering is needed before training an autoencoder. Moreover, the selected error threshold is crucial for the performance.

Apart from error-based methods, recent methods use a deep network to replace non-linear kernel methods of one-class classifiers~\cite{ruff2018,Chalapathy2018}. For instance, the state-of-the-art method Deep SVDD proposed by Ruff et al.~\cite{ruff2018} combines a deep network with an SVDD. Different from end-to-end models, Perera et al.~\cite{perera2018} proposed a deep model to extract one-class features from raw data which are subsequently given into a conventional one-class classifier. However, their method requires a multi-class reference dataset which is difficult to acquire for a given one-class problem. Moreover, the method can only guarantee a tight decision boundary if the reference dataset is highly correlated to the abnormal data. Finally, as the model is not an end-to-end model, the objectives for feature extraction and classification are disconnected.

% this paper
In this paper, we focus on end-to-end models and introduce a generic method for one-class classification using an arbitrary deep neural network as a backbone. The key is to split training samples of the normal class into two subsets, typical normal and atypical normal. Fig.~\ref{fig:examples_typical_atypical} shows examples randomly sampled from these two subsets. By using a binary loss and applying distance constraints on the two subsets, the proposed method enables end-to-end training. Eventually, the output of the network can be interpreted as the probability that a given sample belongs to the normal class.

\section{Proposed method}
\label{sec:proposed_method}
Our basic idea is to split training data into two subsets, namely typical and atypical samples. We call this intra-class splitting (ICS). This enables to use a binary loss during training and to define distance constraints between these subsets. Moreover, tight and closed decision boundaries necessary for one-class classification can be achieved despite training a conventional deep neural network in an end-to-end manner.

Fig.~\ref{fig:network} visualizes the proposed architecture.
%It consists of a feature extraction, a classification and a distance subnetwork. In particular, any conventional deep neural network can form the concatenated feature extraction and classification subnetwork.
A final one-class classifier utilizes an arbitrary deep neural network as a backbone, in which the previous layers can be considered as feature extraction subnetwork and the top layer corresponds to a classification subnetwork.
In contrast to these two subnetworks, the distance subnetwork is only used during training to satisfy some constraints on the latent representations~$\bm{z}$.

\subsection{Intra-Class Splitting}
In a given normal class, not all samples are representative for this class as illustrated in Fig.~\ref{fig:examples_typical_atypical}.
Therefore, it is assumed that a given normal dataset is composed of two disjunct subsets. The first subset consists of typical normal samples which are the most representative for the normal class and correspond to the majority of the given dataset. The second subset is considered to contain atypical normal samples.

An intuitive approach to split a given normal dataset $\mathbb{X}$ is to train a neural network with a bottleneck structure such as an autoencoder.
By using a compression-decompression process, only the most important information of the input data is well maintained.
Accordingly, those samples $\bm{x}$ which are better reconstructed contain more representative features.
Hence, the first step of ICS is to train an autoencoder with all given normal samples using MSE as the objective. 

After training an autoencoder, it is used to acquire the reconstructions of all training samples. Then, the similarity between the original data $\bm{x}$ and the reconstructed data $\hat{\bm{x}}$ is calculated using a predefined similarity metric $\simfunc(\bm{x}, \hat{\bm{x}})$. For example, the structural similarity (SSIM)~\cite{ssim2003} is a possible similarity metric for image data.

Finally, according to the predefined ratio $\rho$, the first $\rho\%$ samples with the lowest similarity scores are considered as atypical normal samples $\mathbb{X}_\mathrm{atypical}$, while the others are considered as typical normal samples $\mathbb{X}_\mathrm{typical}$.

\begin{figure}[t]
	\centering
	\includegraphics[width=0.9\linewidth, trim={0cm 0 0 0}, clip]{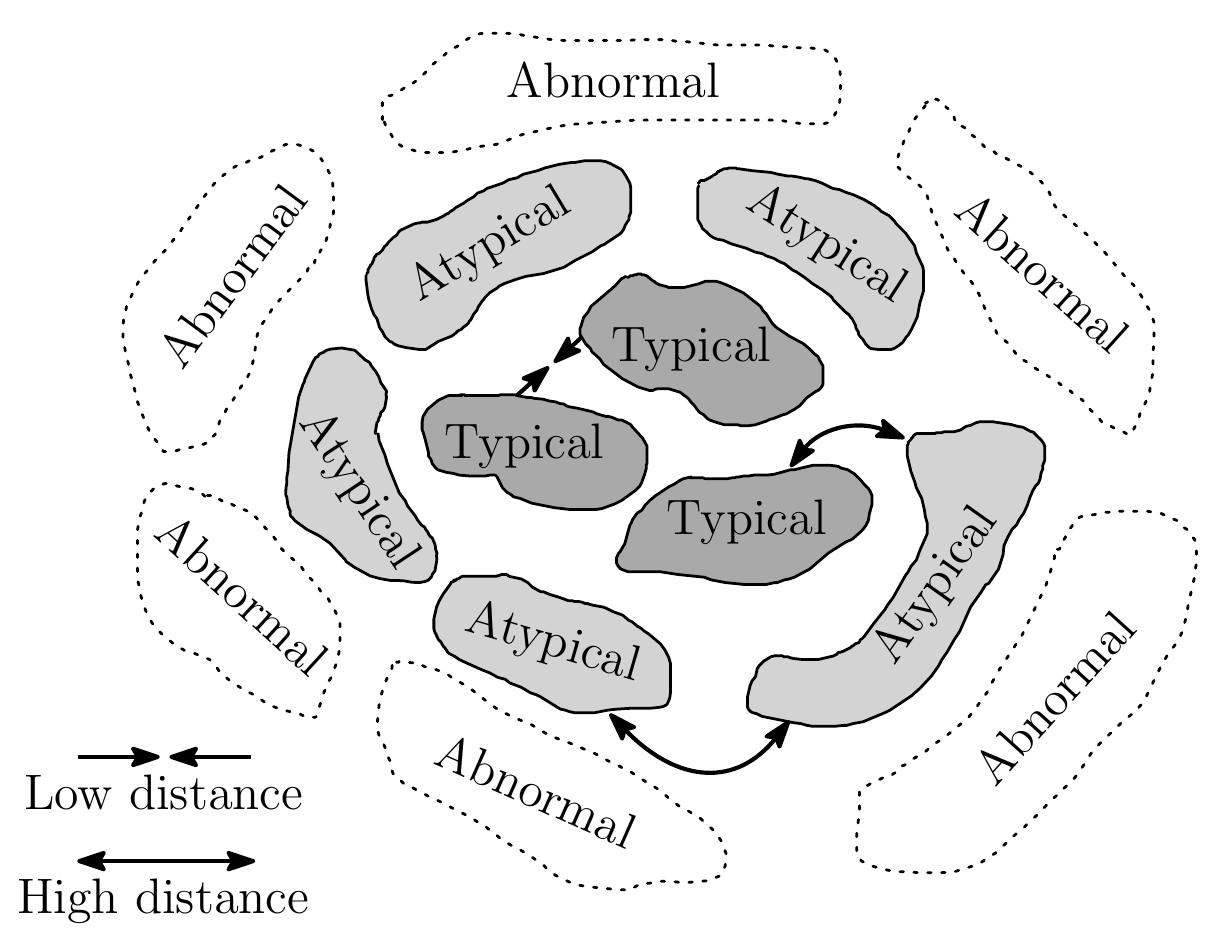}
	\vspace{-0.1cm}
	\caption{Basic idea of intra-class splitting and one-class constraints.}
	\label{fig:venn}
\end{figure}

\subsection{One-Class Constraints}
Based on ICS, three distance constraints on the desired latent representations $\bm{z}$ are defined and visualized in Fig.~\ref{fig:venn}:
\begin{enumerate}[leftmargin=*]
	\item \textbf{Small distances among typical normal samples.} The typical normal samples represent the given normal class compactly. Thus the latent representations of different typical normal samples should indicate similar high-level features. In other words, the latent representations of typical normal samples should have small distances to each other.
	
	\item \textbf{Large distances between typical and atypical normal samples.} Atypical normal samples are assumed to have high-level features more similar to the abnormal samples than those of typical normal samples. Therefore, the latent representations of atypical normal samples should be easily discriminable from those of typical normal samples, i.e. the latent representations of typical and atypical normal samples should have large distances.
	
	\item \textbf{Large distances among atypical normal samples.} The latent representations of atypical normal samples should have large distances among themselves in order to force that typical normal samples are enclosed by atypical samples. This is the key to a tight and closed decision boundary.
\end{enumerate}
Since the term ``distance'' is not restricted to a specific distance metric, we allow a generic, differentiable distance function $\distfunc(\cdot)$ as in~\cite{dosovitskiy2016generating,johnson2016perceptual}. It is modeled by the distance subnetwork in Fig.~\ref{fig:network}. $\distfunc(\cdot)$ takes two inputs and calculates a scalar value normalized in the range of $[0, 1]$ as the distance between these two inputs. According to the above constraints, $\distfunc(\cdot)$ is learned under the following criteria:
\begin{equation}
\distfunc(\bm{z}_i,\bm{z}_{j\neq i})
\overset{!}{=}
\begin{cases}
0, \bm{z}_i \text{ and } \bm{z}_j \text{ typical}\\
1, \text{else}
\end{cases}
\end{equation}

\subsection{Training}
\label{subsec:training}
After ICS, the three subnetworks from Fig.~\ref{fig:network} are jointly trained with typical and atypical normal data. For each of the three desired constraints on the latent representations $\bm{z}$, a loss is defined. Then, the network is trained with the three different losses step by step for a fixed number of iterations.

During the first step, the network is trained with a batch of typical normal samples to minimize the distance between their latent representations by minimizing the \emph{closeness loss}
\begin{equation}
\mathcal{L}_\mathrm{cls} = -\frac{1}{B}\sum_{i=1}^{B}\log\left(1-\distfunc(\bm{z}_{\mathrm{typ}, i},\bm{z}_{\mathrm{typ}, j\neq i})\right)~,
\end{equation}
where $\bm{z}_{\mathrm{typ}, j\neq i}$ and $\bm{z}_{\mathrm{typ}, i}$ are the latent representations of two different typical normal samples.

Second, assigning the label ``0'' to typical normal samples and the label ``1'' to atypical normal samples enables to train the network with a binary cross-entropy loss. We call this loss \emph{intra-class loss} $\mathcal{L}_\mathrm{ic}$:
\begin{equation}
\mathcal{L}_\mathrm{ic} = -\frac{1}{B}\sum_{i=1}^{B}\left[y_i\log\hat{y}_i + (1-y_i)\log(1-\hat{y}_i)\right]~,
\end{equation}
where $y_i$ is the label for a given sample $\bm{x}_i$ and $\hat{y}_i = f(\bm{x}_i)$ is the label of $\bm{x}_i$ predicted by the classification subnetwork. Thereby, the function $f(\cdot)$ is realized by the one-class classifier consisting of the feature extraction and classification subnetwork.
This loss implicitly maximizes the distances between the latent representations of typical and atypical normal samples.

Third, the network is trained with only atypical normal samples to maximize the distances among latent representations of atypical normal samples. This is done by minimizing the \emph{dispersion loss}
\begin{equation}
\mathcal{L}_\mathrm{disp} = -\frac{1}{B}\sum_{i=1}^{B}\log \distfunc(\bm{z}_{\mathrm{atyp}, i},\bm{z}_{\mathrm{atyp}, j\neq i})~,
\end{equation}
where $\bm{z}_{\mathrm{atyp}, i}$ and $\bm{z}_{\mathrm{atyp}, j\neq i}$ are the latent representations of two different atypical normal samples.

\begin{table*}[htb]
	\centering
	\caption{AUC (standard deviation) in \%.}
	\vspace{0.1cm}
	\label{tab:auc_vs_baseline}
	\scriptsize
	\def\arraystretch{1.0}
	\setlength{\tabcolsep}{0.25cm}
	\begin{tabular}{@{}lcccccccc@{}}
		\toprule
		\textbf{Normal Class} & \textbf{OCSVM} & \textbf{IF} & \textbf{ImageNet} & \textbf{SSIM} & \textbf{DSVDD} & \textbf{NaiveNN} & \textbf{NNwICS} & \textbf{Ours} \\
		\midrule
		Digit 0    	& 98.2 {\scriptsize($\pm$0.0)} & 96.2 {\scriptsize($\pm$0.3)} & 71.1 {\scriptsize($\pm$0.0)} & 98.7 {\scriptsize($\pm$0.0)} & 98.0 {\scriptsize($\pm$0.7)} & 96.8 {\scriptsize($\pm$0.5)} & 96.9 {\scriptsize($\pm$0.1)} & \textbf{98.9 {\scriptsize($\pm$0.0)}} \\
		Digit 1    	& 99.2 {\scriptsize($\pm$0.0)} & 99.4 {\scriptsize($\pm$0.0)} & 88.9 {\scriptsize($\pm$0.0)} & \textbf{99.8 {\scriptsize($\pm$0.0)}} & 99.7 {\scriptsize($\pm$0.1)} & 74.8 {\scriptsize($\pm$6.0)} & 98.7 {\scriptsize($\pm$0.3)} & \textbf{99.8 {\scriptsize($\pm$0.1)}} \\
		Digit 2    	& 82.1 {\scriptsize($\pm$0.0)} & 73.0 {\scriptsize($\pm$2.3)} & 58.5 {\scriptsize($\pm$0.0)} & 82.6 {\scriptsize($\pm$0.0)} & \textbf{91.7 {\scriptsize($\pm$0.8)}} & 67.5 {\scriptsize($\pm$5.2)} & 85.4 {\scriptsize($\pm$2.6)} & \textbf{91.7 {\scriptsize($\pm$1.8)}} \\
		Digit 3    	& 86.1 {\scriptsize($\pm$0.0)} & 82.6 {\scriptsize($\pm$0.5)} & 63.2 {\scriptsize($\pm$0.0)} & 90.6 {\scriptsize($\pm$0.0)} & 91.9 {\scriptsize($\pm$1.5)} & 67.1 {\scriptsize($\pm$2.8)} & 95.1 {\scriptsize($\pm$0.3)} & \textbf{96.6 {\scriptsize($\pm$0.2)}} \\
		Digit 4    	& 94.8 {\scriptsize($\pm$0.0)} & 87.9 {\scriptsize($\pm$0.4)} & 70.6 {\scriptsize($\pm$0.0)} & 76.6 {\scriptsize($\pm$0.0)} & \textbf{94.9 {\scriptsize($\pm$0.8)}} & 94.1 {\scriptsize($\pm$0.5)} & 77.5 {\scriptsize($\pm$0.6)} & 86.5 {\scriptsize($\pm$1.3)} \\
		Digit 5    	& 77.4 {\scriptsize($\pm$0.0)} & 73.4 {\scriptsize($\pm$0.8)} & 60.7 {\scriptsize($\pm$0.0)} & \textbf{92.3 {\scriptsize($\pm$0.0)}} & 88.5 {\scriptsize($\pm$0.9)} & 55.8 {\scriptsize($\pm$4.2)} & 85.5 {\scriptsize($\pm$0.7)} & 88.9 {\scriptsize($\pm$0.0)} \\
		Digit 6    	& 94.8 {\scriptsize($\pm$0.0)} & 85.8 {\scriptsize($\pm$0.9)} & 67.6 {\scriptsize($\pm$0.0)} & 96.6 {\scriptsize($\pm$0.0)} & 98.3 {\scriptsize($\pm$0.5)} & 89.7 {\scriptsize($\pm$0.2)} & 96.3 {\scriptsize($\pm$0.0)} & \textbf{98.8 {\scriptsize($\pm$0.2)}} \\
		Digit 7    	& 93.4 {\scriptsize($\pm$0.0)} & 91.4 {\scriptsize($\pm$0.5)} & 71.0 {\scriptsize($\pm$0.0)} & 96.0 {\scriptsize($\pm$0.0)} & 94.6 {\scriptsize($\pm$0.9)} & 68.2 {\scriptsize($\pm$7.3)} & 94.0 {\scriptsize($\pm$0.0)} & \textbf{96.1 {\scriptsize($\pm$0.2)}} \\
		Digit 8    	& 90.2 {\scriptsize($\pm$0.0)} & 73.9 {\scriptsize($\pm$1.1)} & 64.0 {\scriptsize($\pm$0.0)} & 80.2 {\scriptsize($\pm$0.0)} & 93.9 {\scriptsize($\pm$1.6)} & 77.7 {\scriptsize($\pm$9.2)} & 91.0 {\scriptsize($\pm$0.3)} & \textbf{95.0 {\scriptsize($\pm$0.2)}} \\
		Digit 9    	& 92.8 {\scriptsize($\pm$0.0)} & 87.5 {\scriptsize($\pm$0.1)} & 71.8 {\scriptsize($\pm$0.0)} & 79.5 {\scriptsize($\pm$0.0)} & \textbf{96.5 {\scriptsize($\pm$0.3)}} & 81.8 {\scriptsize($\pm$0.7)} & 87.4 {\scriptsize($\pm$0.1)} & 90.0 {\scriptsize($\pm$0.4)} \\
		\midrule                                                                                                   
		T-shirt    	& 86.1 {\scriptsize($\pm$0.0)} & 86.8 {\scriptsize($\pm$0.6)} & 58.1 {\scriptsize($\pm$0.0)} & 83.7 {\scriptsize($\pm$0.0)} & 79.1 {\scriptsize($\pm$1.5)}  & 62.9 {\scriptsize($\pm$0.9)} & 85.1 {\scriptsize($\pm$1.7)} & \textbf{88.3 {\scriptsize($\pm$1.2)}} \\
		Trouser    	& 93.9 {\scriptsize($\pm$0.0)} & 97.7 {\scriptsize($\pm$0.1)} & 75.4 {\scriptsize($\pm$0.0)} & 98.5 {\scriptsize($\pm$0.0)} & 94.0 {\scriptsize($\pm$1.3)}  & 65.6 {\scriptsize($\pm$4.5)} & 94.6 {\scriptsize($\pm$0.1)} & \textbf{98.9 {\scriptsize($\pm$0.2)}} \\
		Pullover   	& 85.6 {\scriptsize($\pm$0.0)} & 87.1 {\scriptsize($\pm$0.3)} & 58.1 {\scriptsize($\pm$0.0)} & 87.2 {\scriptsize($\pm$0.0)} & 83.0 {\scriptsize($\pm$1.4)}  & 73.6 {\scriptsize($\pm$0.9)} & 82.6 {\scriptsize($\pm$1.2)} & \textbf{88.2 {\scriptsize($\pm$0.4)}} \\
		Dress      	& 85.9 {\scriptsize($\pm$0.0)} & 90.1 {\scriptsize($\pm$0.7)} & 60.1 {\scriptsize($\pm$0.0)} & 89.2 {\scriptsize($\pm$0.0)} & 82.9 {\scriptsize($\pm$1.9)}  & 70.0 {\scriptsize($\pm$1.7)} & 89.1 {\scriptsize($\pm$0.1)} & \textbf{92.1 {\scriptsize($\pm$2.2)}} \\
		Coat       	& 84.6 {\scriptsize($\pm$0.0)} & 89.8 {\scriptsize($\pm$0.4)} & 58.3 {\scriptsize($\pm$0.0)} & 87.3 {\scriptsize($\pm$0.0)} & 87.0 {\scriptsize($\pm$0.5)}  & 80.8 {\scriptsize($\pm$3.9)} & 85.8 {\scriptsize($\pm$0.2)} & \textbf{90.2 {\scriptsize($\pm$0.0)}} \\
		Sandal     	& 81.3 {\scriptsize($\pm$0.0)} & 88.7 {\scriptsize($\pm$0.2)} & 69.2 {\scriptsize($\pm$0.0)} & 85.2 {\scriptsize($\pm$0.0)} & 80.3 {\scriptsize($\pm$4.6)}  & 64.0 {\scriptsize($\pm$9.4)} & 85.5 {\scriptsize($\pm$0.0)} & \textbf{89.4 {\scriptsize($\pm$1.4)}} \\
		Shirt      	& 78.6 {\scriptsize($\pm$0.0)} & \textbf{79.7 {\scriptsize($\pm$0.9)}} & 57.3 {\scriptsize($\pm$0.0)} & 75.3 {\scriptsize($\pm$0.0)} & 74.9 {\scriptsize($\pm$1.3)}  & 71.8 {\scriptsize($\pm$1.3)} & 75.6 {\scriptsize($\pm$0.4)} & 78.3 {\scriptsize($\pm$0.6)} \\
		Sneaker    	& 97.6 {\scriptsize($\pm$0.0)} & 98.0 {\scriptsize($\pm$0.1)} & 75.5 {\scriptsize($\pm$0.0)} & 97.8 {\scriptsize($\pm$0.0)} & 94.2 {\scriptsize($\pm$2.1)}  & 92.0 {\scriptsize($\pm$3.2)} & 94.9 {\scriptsize($\pm$0.1)} & \textbf{98.3 {\scriptsize($\pm$0.2)}} \\
		Bag        	& 79.5 {\scriptsize($\pm$0.0)} & 88.3 {\scriptsize($\pm$0.6)} & 61.9 {\scriptsize($\pm$0.0)} & 81.6 {\scriptsize($\pm$0.0)} & 79.1 {\scriptsize($\pm$4.5)}  & 72.9 {\scriptsize($\pm$8.7)} & 82.0 {\scriptsize($\pm$0.3)} & \textbf{88.6 {\scriptsize($\pm$2.3)}} \\
		Ankle boot 	& 97.8 {\scriptsize($\pm$0.0)} & 97.9 {\scriptsize($\pm$0.1)} & 78.3 {\scriptsize($\pm$0.0)} & 98.4 {\scriptsize($\pm$0.0)} & 93.2 {\scriptsize($\pm$2.4)}  & 90.7 {\scriptsize($\pm$0.1)} & 94.9 {\scriptsize($\pm$0.3)} & \textbf{98.5 {\scriptsize($\pm$0.1)}} \\
		\midrule	    	                                                                                       
		Airplane   	& 61.9 {\scriptsize($\pm$0.0)} & 66.7 {\scriptsize($\pm$1.3)} & 53.3 {\scriptsize($\pm$0.0)} & 75.6 {\scriptsize($\pm$0.0)} & 61.7 {\scriptsize($\pm$4.1)} & 63.8 {\scriptsize($\pm$4.5)} & 62.7 {\scriptsize($\pm$1.8)} & \textbf{76.8 {\scriptsize($\pm$3.2)}} \\
		Automobile 	& 38.5 {\scriptsize($\pm$0.0)} & 43.6 {\scriptsize($\pm$1.3)} & 53.6 {\scriptsize($\pm$0.0)} & 43.5 {\scriptsize($\pm$0.0)} & 65.9 {\scriptsize($\pm$2.1)} & 52.7 {\scriptsize($\pm$0.9)} & 63.2 {\scriptsize($\pm$0.8)} & \textbf{71.3 {\scriptsize($\pm$0.2)}} \\
		Bird       	& 60.6 {\scriptsize($\pm$0.0)} & 59.1 {\scriptsize($\pm$0.3)} & 51.9 {\scriptsize($\pm$0.0)} & 61.1 {\scriptsize($\pm$0.0)} & 50.8 {\scriptsize($\pm$0.8)} & 47.8 {\scriptsize($\pm$0.4)} & 57.6 {\scriptsize($\pm$0.4)} & \textbf{63.0 {\scriptsize($\pm$0.8)}} \\
		Cat        	& 49.4 {\scriptsize($\pm$0.0)} & 50.3 {\scriptsize($\pm$0.5)} & 50.8 {\scriptsize($\pm$0.0)} & 48.6 {\scriptsize($\pm$0.0)} & 59.1 {\scriptsize($\pm$1.4)} & 50.2 {\scriptsize($\pm$4.3)} & 58.0 {\scriptsize($\pm$0.2)} & \textbf{60.1 {\scriptsize($\pm$3.4)}} \\
		Deer       	& 71.3 {\scriptsize($\pm$0.0)} & 74.4 {\scriptsize($\pm$0.2)} & 55.8 {\scriptsize($\pm$0.0)} & 63.5 {\scriptsize($\pm$0.0)} & 60.9 {\scriptsize($\pm$1.1)} & 65.1 {\scriptsize($\pm$1.8)} & 61.9 {\scriptsize($\pm$0.1)} & \textbf{74.9 {\scriptsize($\pm$0.9)}} \\
		Dog        	& 52.0 {\scriptsize($\pm$0.0)} & 51.4 {\scriptsize($\pm$0.3)} & 52.6 {\scriptsize($\pm$0.0)} & 62.1 {\scriptsize($\pm$0.0)} & 65.7 {\scriptsize($\pm$2.5)} & 53.3 {\scriptsize($\pm$0.7)} & 65.7 {\scriptsize($\pm$0.2)} & \textbf{66.0 {\scriptsize($\pm$1.1)}} \\
		Frog       	& 63.8 {\scriptsize($\pm$0.0)} & 71.1 {\scriptsize($\pm$0.5)} & 54.6 {\scriptsize($\pm$0.0)} & 44.5 {\scriptsize($\pm$0.0)} & 67.7 {\scriptsize($\pm$2.6)} & 41.1 {\scriptsize($\pm$3.4)} & 64.2 {\scriptsize($\pm$2.5)} & \textbf{71.6 {\scriptsize($\pm$0.8)}} \\
		Horse      	& 48.2 {\scriptsize($\pm$0.0)} & 53.6 {\scriptsize($\pm$0.3)} & 51.3 {\scriptsize($\pm$0.0)} & 47.2 {\scriptsize($\pm$0.0)} & \textbf{67.3 {\scriptsize($\pm$0.9)}} & 51.5 {\scriptsize($\pm$0.4)} & 62.4 {\scriptsize($\pm$2.0)} & 64.1 {\scriptsize($\pm$1.6)} \\
		Ship       	& 63.7 {\scriptsize($\pm$0.0)} & 69.4 {\scriptsize($\pm$0.6)} & 57.1 {\scriptsize($\pm$0.0)} & 76.8 {\scriptsize($\pm$0.0)} & 75.9 {\scriptsize($\pm$1.2)} & 45.8 {\scriptsize($\pm$5.0)} & 73.9 {\scriptsize($\pm$1.1)} & \textbf{78.9 {\scriptsize($\pm$0.5)}} \\
		Truck		& 48.8 {\scriptsize($\pm$0.0)} & 53.9 {\scriptsize($\pm$1.0)} & 56.0 {\scriptsize($\pm$0.0)} & 40.7 {\scriptsize($\pm$0.0)} & \textbf{73.1 {\scriptsize($\pm$1.2)}} & 53.5 {\scriptsize($\pm$3.2)} & 55.6 {\scriptsize($\pm$2.2)} & 66.0 {\scriptsize($\pm$2.5)} \\
		\bottomrule
	\end{tabular}
\end{table*}

\section{Experiments}
\label{sec:experiments}

\subsection{Setup}
% datasets
The proposed method was evaluated on three benchmark image datasets MNIST~\cite{lecun1998gradient}, Fashion-MNIST~\cite{fashionmnist2015} and CIFAR-10~\cite{krizhevsky2009learning}. All three datasets have ten different classes. Per dataset, one class was taken as the normal class and the remaining nine classes were considered as abnormal classes. Accordingly, the training set sizes were $n\approx 6000$ for MNIST, $n=6000$ for Fashion-MNIST and $n=5000$ for CIFAR-10. The test set was composed of 1000 normal samples and 9000 abnormal samples. Finally, AUC~\cite{auc1997} was used as performance metric.

% baseline models
According to the literature, only few prior work proposed state-of-the-art one-class classifiers. In this work, the following conventional and deep learning based models were selected as baseline models: \emph{i) OCSVM~\cite{ocsvm2001}} using $\nu = 0.15$ and an RBF kernel with kernel size $\gamma = \frac{1}{\#\mathrm{features}}$; \emph{ii) Isolation Forest (IF)~\cite{if2008}}; \emph{iii) ImageNet + OCSVM}: Features extracted by a VGG19~\cite{Simonyan15} pretrained on ImageNet~\cite{Russakovsky2015} were used as the input for a OCSVM; \emph{iv) Deep SVDD (DSVDD)~\cite{ruff2018}}; \emph{v) Error based classifier} (SSIM) which directly took the SSIM between the reconstructions and the original data as the classification score. OCSVM, IF and DSVDD shared the settings from~\cite{ruff2018}.
In addition, the following variants of the proposed method were considered as baseline models: \emph{vi) Naive neural network without ICS (NaiveNN)}: The network with the same architecture as the proposed method was trained without the distance subnetwork or ICS; \emph{vii) Neural network with ICS but without one-class constraints (NNwDS)}: The normal dataset was split into typical and atypical subsets. After assigning two different labels to the subsets, the network was trained with these two subsets but without any constraints on the latent representations.

% network architecture
The concrete architecture of the autoencoder for ICS is arbitrary. In this work, the encoder shared a similar structure with AlexNet~\cite{alexnet2012} except that all dense layers were replaced by one convolutional layer. The decoder had a symmetrical structure to the encoder, which utilized transposed convolutional layers for upsampling.
The base architecture for the proposed method was AlexNet. The feature extraction subnetwork in Fig.~\ref{fig:network} was composed of the layers from the input layer to the second last layer of the AlexNet. Its output layer was considered as the classification subnetwork. The distance subnetwork was composed of one subtraction layer and one dense layer. In particular, the subtraction layer calculated the pixel-wise difference $\Delta \bm{z}=\bm{z}_i-\bm{z}_j$ which was subsequently mapped to a scalar value by the dense layer. Note that these two subnetworks can be replaced by any other deeper networks.

% hyperparameters
The proposed model was implemented with TensorFlow~\cite{abadi2016tensorflow}. We used SSIM as similarity metric for ICS. Furthermore, the ratio $\rho$ for choosing atypical normal samples was set to 10 and the number of training iterations was 10000. Finally, the training mini-batch size was 64 and L2-regularization was used for every convolutional layer with a decay of $10^{-6}$.

\subsection{\vspace{-0.2cm}Results and Discussion}
Table~\ref{tab:auc_vs_baseline} shows the resulting AUCs in percent averaged over five different seeds for the initialization of the network. Compared to the baseline models, the proposed method performed best in 23 of 30 cases. Moreover, our method showed a better performance than the baseline models especially for the natural image dataset CIFAR-10. For example, the proposed method outperformed the recent state-of-the-art method DSVDD in 8 of 10 cases on CIFAR-10 with an average improvement of more than 11.4\%.

Although all methods performed similarly on the trivial datasets, the proposed method still showed improved performance. For instance, on MNIST, our method achieved 1.4\% improvement over DSVDD and over 6.5\% improvement over OCSVM, IF, ImageNet and SSIM.
However, for some normal classes, e.g. digit 4, the proposed method underperformed state-of-the-art methods due to the non-optimized ratio $\rho$.
%In these cases, finetuning $\rho$ can help to further improve the performance in these cases.

Considering the variants of the proposed method, the NaiveNN performed worst as expected, because it tends to map all points from the original data space to an identical label, making a correct classification challenging. This situation is tolerable in simple datasets. However, the NaiveNN cannot be used at all on the more complex dataset CIFAR-10. In contrast, NNwICS, a naive neural network with ICS, achieved higher AUCs and was comparable to or outperformed the other baseline models. In conclusion, the integration of ICS into neural networks can enhance the performance for one-class classification problems.

%Considering the variants of the proposed method, the NaiveNN performed worst as expected, because it tends to map all points from the original data space to an identical label, making a correct classification challenging. Although this situation is tolerable in simple datasets, the NaiveNN cannot be used at all on the more complex dataset CIFAR-10. In contrast, NNwICS, a naive neural network with ICS, achieved higher AUCs and was comparable to or outperformed the other baseline models. In conclusion, the integration of ICS into neural networks can enhance the performance for one-class classification problems.

%Indeed, the proposed method underperformed state-of-the-art methods for some normal classes, e.g. digit 4, 5 and 9. Indeed, finetuning $\rho$ can help to further improve the performance in these cases.

%Indeed, for some normal classes (e.g. digit 4, 5 and 9), the proposed method underperformed state-of-the-art methods. In these cases, finetuning $\rho$ can help to further improve the performance in these cases.

\begin{figure}
	\centering
	\includegraphics[width=\linewidth]{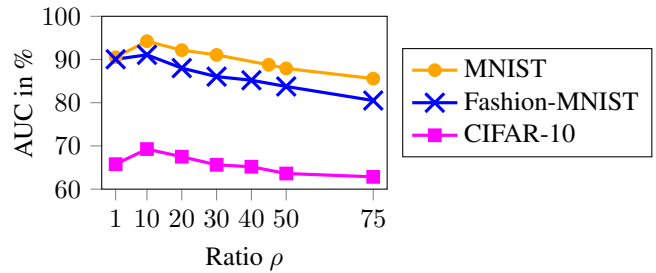}
	\vspace{-0.7cm}
	\caption{AUC over ratio $\rho$.}
	\label{fig:AUC_vs_thr}
	\vspace{-0.3cm}
\end{figure}

Eventually, the proposed method was evaluated with different ratios $\rho$ to judge its sensitivity. Fig.~\ref{fig:AUC_vs_thr} shows AUCs averaged over ten classes and four different initialization seeds for each dataset depending on the ratio $\rho$.
The results indicate that each dataset has one optimal $\rho$ which is about 10\%.
%In particular, Table~\ref{tab:auc_vs_baseline} also shows each experiment has an optimal $\rho$.
%This also explains why the proposed method underperformed state-of-the-art methods for some normal classes, e.g. digit 4, 5 and 9. Indeed, finetuning $\rho$ can help to further improve the performance in these cases.
%The insight into the dependency on $\rho$ can also explain why the proposed method underperformed state-of-the-art methods for some normal classes, e.g. digit 4, 5 and 9. Indeed, finetuning $\rho$ can help to further improve the performance in these cases.
By choosing a smaller or greater value for $\rho$, the AUC is worse.

\section{\vspace{-0.2cm}Conclusion}
\label{sec:conclusion}
We proposed a novel method for one-class classification using deep learning. By splitting given normal data into typical and atypical normal subsets, it allows to introduce a binary loss and additional constraints which enable an end-to-end training of standard deep neural networks.
The proposed method was evaluated in various experiments on image datasets. It showed a distinct improvement over state-of-the-art approaches to one-class classification in average, especially for the complex dataset CIFAR-10.
Future implications of this paper may include the extension of the proposed method to larger network architectures and more complex datasets. Moreover, the proposed method may be transferred to the field of open set recognition. Finally, we work on mathematical proofs for the significance of intra-class splitting.

% References.
% -------------------------------------------------------------------------
\renewcommand{\baselinestretch}{0.93}\selectfont
\let\OLDthebibliography\thebibliography
\renewcommand\thebibliography[1]{
	\OLDthebibliography{#1}
	\setlength{\parskip}{0cm}
	\setlength{\itemsep}{0.16cm}
}
\bibliographystyle{IEEEbib}
\bibliography{references}	
\end{document}